\documentclass[letterpaper]{article} 
\usepackage{aaai2026}  
\usepackage{times}  
\usepackage{helvet}  
\usepackage{courier}  
\usepackage[hyphens]{url}  
\usepackage{graphicx} 
\usepackage{natbib}  
\usepackage{caption} 
\usepackage{algorithm}
\usepackage{algorithmic}
\usepackage{amssymb} 
\usepackage{amsmath}
\usepackage{booktabs}
\usepackage{multirow}

\usepackage{newfloat}
\usepackage{listings}
\DeclareCaptionStyle{ruled}{labelfont=normalfont,labelsep=colon,strut=off} 
\floatstyle{ruled}
\newfloat{listing}{tb}{lst}{}
\floatname{listing}{Listing}
\title{Spectral Property-Driven Data Augmentation for Hyperspectral Single‑Source Domain Generalization}
\author{
    Taiqin Chen\textsuperscript{\rm 1, \rm 2},
    Yifeng Wang\textsuperscript{\rm 3},
    Xiaochen Feng\textsuperscript{\rm 1},
    Zhilin Zhu,\textsuperscript{\rm 1, \rm 2},
    Hao Sha\textsuperscript{\rm 1},
    Yingjian Li,\textsuperscript{\rm 2},
    Yongbing Zhang\textsuperscript{\rm 1}\thanks{Corresponding author: Yongbing Zhang}
}
\affiliations{
    \textsuperscript{\rm 1}School of Computer Science and Technology, Harbin Institute of Technology (Shenzhen), Shenzhen, China\\
    \textsuperscript{\rm 2}Pengcheng Laboratory, Shenzhen, China\\
    \textsuperscript{\rm 3}Department of Computer Science and Technology, Tsinghua University, Beijing, China\\

    chentq@stu.hit.edu.cn, ybzhang08@hit.edu.cn
}

\usepackage{bibentry}

\begin{document}

\maketitle

\begin{abstract}
While hyperspectral images (HSI) benefit from numerous spectral channels that provide rich information for classification, the increased dimensionality and sensor variability make them more sensitive to distributional discrepancies across domains, which in turn can affect classification performance. To tackle this issue, hyperspectral single-source domain generalization (SDG) typically employs data augmentation to simulate potential domain shifts and enhance model robustness under the condition of single-source domain training data availability. However, blind augmentation may produce samples misaligned with real-world scenarios, while excessive emphasis on realism can suppress diversity, highlighting a tradeoff between realism and diversity that limits generalization to target domains. To address this challenge, we propose a spectral property-driven data augmentation (SPDDA) that explicitly accounts for the inherent properties of HSI, namely the device-dependent variation in the number of spectral channels and the mixing of adjacent channels. Specifically, SPDDA employs a spectral diversity module that resamples data from the source domain along the spectral dimension to generate samples with varying spectral channels, and constructs a channel-wise adaptive spectral mixer by modeling inter-channel similarity, thereby avoiding fixed augmentation patterns. To further enhance the realism of the augmented samples, we propose a spatial-spectral co-optimization mechanism, which jointly optimizes a spatial fidelity constraint and a spectral continuity self-constraint. Moreover, the weight of the spectral self-constraint is adaptively adjusted based on the spatial counterpart, thus preventing over-smoothing in the spectral dimension and preserving spatial structure. Extensive experiments conducted on three remote sensing benchmarks demonstrate that SPDDA outperforms state-of-the-art methods.
\end{abstract}

\begin{links}
    \link{Code}{https://github.com/hnsytq/SPDDA}
\end{links}

\section{Introduction}

Hyperspectral image (HSI) is capable of accurately depicting the intrinsic properties of matter, which is facilitated by its simultaneous representation of both spectral and spatial information of the imaging scene. Compared with RGB images, HSI shows better recognition performance with its richer spectral depicting ability, and thus has a broad application prospect in the fields of remote sensing, medical diagnosis, etc~\cite{HSI_review}.

As a fundamental task in HSI-based analysis, hyperspectral image classification (HSIC) for remote sensing~\cite{HSIC_review} achieves pixel-level land-cover identification in regions of interest. With the rapid development of deep learning~\cite{HSIC_mamba, yuan2018unsupervised}, the performance of HSIC has been significantly improved. Deep learning-based methods typically leverage a large annotated dataset to learn mappings from HSI to categorical labels~\cite{DG_review_1}, under the assumption that test samples are independent and identically distributed with the training data, where test samples are referred to as the target domain (TD) and the training data is referred to as the source domain (SD). However, in practice, substantial discrepancies often exist between the training and testing data distributions due to multiple factors such as illumination and weather~\cite{yang2021miniaturization}. This domain disparity results in a dramatic degradation of classification performance~\cite{DG_review_2}.

To tackle this issue, domain generalization (DG) is proposed to learn domain-invariant features with robustness and transferability in the training phase~\cite{DG_first}, thereby enhancing classification performance on the unseen TD. The classical setting of DG assumes that multi-source domain data with annotations can be accessed during classifier training~\cite{PACS}. However, obtaining labeled data from multiple domains is often impractical due to data privacy and expensive annotation costs in real-world remote sensing scenarios. Therefore, a more challenging setting that is single source domain generalization (SDG)~\cite{SDENet} is more suitable for HSIC, where only samples from a single SD is available.

Since single-source samples are insufficient to support the learning of domain-invariant features, existing methods typically employ data augmentation to simulate distributional shift between different domains in SDG. In brief, these methods typically apply either standard augmentations~\cite{Advst} or trainable generative models~\cite{SDG_21} to introduce random perturbations into SD, thereby generating diverse synthetic samples, referred to as the extended domain (ED). However, HSI depicts real-world scenes, which is contrasty to various stylistic variants of RGB images such as cartoons and skeletons. The mentioned blind data augmentation may generate unrealistic samples, thus biasing the classifier away from its intended prediction mechanism. To tackle this challenge, pioneering studies~\cite{FDGNet, S2AMSnet} have introduced semantic constraints to ensure the reliability of ED during augmentation process. Unfortunately, such constraints inevitably result in similarity between ED and SD, thereby limiting the augmentation diversity. Overall, existing hyperspectral SDG methods suffer from a diversity-realism tradeoff, which constrains the performance of classifiers in unknown TD.


To tackle this challenge, we explicitly account for inherent properties of HSI during augmentation procedure. Specifically, HSI exhibits substantial variability across different scenes and imaging devices~\cite{ODSI}. The spectral sensing range and the number of spectral channels of HSI are varied, which depend on the employed imaging devices or specific application requirements, referred to as spectral heterogeneity~\cite{spectral_vary}. Moreover, the acquired HSI may be aliased due to overlapping sensor spectral responses or atmospheric scattering interference~\cite{spectral_mixing, LMM}, causing adjacent channels in HSI to be mixed or single pixel mixing the spectral responses of multiple matters. The above properties lead to potentially huge distribution shift even for HSIs describing the same scene. In other words, introducing random perturbations into SD grounded in these inherent properties can avoid imposing strict constraints between ED and SD, thereby preserving diversity and realism simultaneously. 



Based on the above analysis, we propose a spectral property-driven data augmentation (SPDDA) to mitigate the tradeoff between diversity and realism in hyperspectral SDG. Specifically, we introduce a spectral diversity module (SDM) for constructing a channel mask and a channel-wise adaptive spectral mixer (CASM). The channel mask is utilized to mask channels with less semantic information, thereby generating HSIs with arbitrary number of channels, corresponding to spectral heterogeneity. CASM then adaptively mixes channels of the generated HSI, whose mixing kernel size and weights are tailored for each channel based on inter-channel similarity. Furthermore, we develop a spatial-spectral co-optimization mechanism (SSCOM) to ensure reliability of the generation, which is comprised of a spatial fidelity constraint and a spectral continuity self-constraint. To alleviate overly smoothening in the spectral dimension, the weight of the spectral self-constraint is dynamically adjusted based on the spatial constraint. Briefly, our contributions are as follow:

\begin{itemize}
    \item We propose SPDDA, a novel augmentation framework that balances the tradeoff between diversity and realism by leveraging the inherent properties of HSI.
    \item We design SDM, which masks a variable number of channels based on semantic information and constructs the CASM to avoid fixed perturbation patterns, thus enhancing the diversity of augmented samples. The procedure is supervised by the SSCOM, a dynamic joint optimization mechanism that constrains across both spatial and spectral dimensions.  
    \item We conduct extensive experiments on three widely-used hyperspectral benchmarks, demonstrating that the proposed method achieves state-of-the-art (SOTA) performance compared to existing SDG methods.

\end{itemize}

\section{Related Work}

\subsection{Domain Generalization}
Domain generalization aims to mitigate the performance degradation caused by the domain shifts, which can be categorized into multi-source domain generalization and single-source domain generalization (SDG). The former intends to learn domain-invariant representations to enhance classifier performance in the unseen target domain (TD) by utilizing labeled samples from multiple source domain during training~\cite{MDG_GNN}. In contrast, SDG merely accesses labeled samples from a single-source domain. Hence, existing methods typically inject perturbation into samples to simulate the domain shifts and obtain diverse extended data. For example, PDEN~\cite{PDEN} proposes a progressive domain expansion network to simulate photometric and geometric transforms in TD. AdvST~\cite{Advst} introduces an adversarial framework that learns parameters of composed standard data augmentations as semantics transformations to generate challenging and diverse samples. UniFreqSDG~\cite{UniFreqSDG} learns adaptive low‑frequency perturbations across multiple feature levels and combines a domain‑divergence inducement loss to enhance analysis performance in TD. PEER~\cite{PEER} utilizes a proxy model in place of the classifier to learn from augmented data and maximizes their mutual information to mitigate effects of distortion. However, directly applying these methods to hyperspectral SDG often fails to effectively exploit spectral characteristics.


\subsection{Hyperspectral Single Source Domain Generation}
SDENet~\cite{SDENet} is the first to apply SDG to hyperspectral image classification to tackle the performance decline caused by domain shifts in cross-scene tasks. SDENet proposes a hyperspectral-tailored generator that comprises a spatial-spectral model and a morph model to obtain augmented samples. SFT~\cite{SFT} proposes a Fourier transformation that learns dynamic attention maps in the frequency domain of source-domain samples to enhance diversity of the augmented samples. However, these methods overemphasize diversity, which may result in generated data deviating from the practical distribution and introducing bias into classifier optimization. To tack this issue, FDGNet~\cite{FDGNet} incorporates data geometry into the generation process to preserve authenticity, enforcing augmented samples to retain low-dimensional structures similar to those of the source domain. S2AMSNet~\cite{S2AMSnet} introduces a mutual information regulation to prevent random perturbation from corrupting the semantic information of samples. However, such a constraint inevitably compromises diversity.

\renewcommand{\dblfloatpagefraction}{.9}
\begin{figure*}
    \centering
    \includegraphics[width=0.8\linewidth]{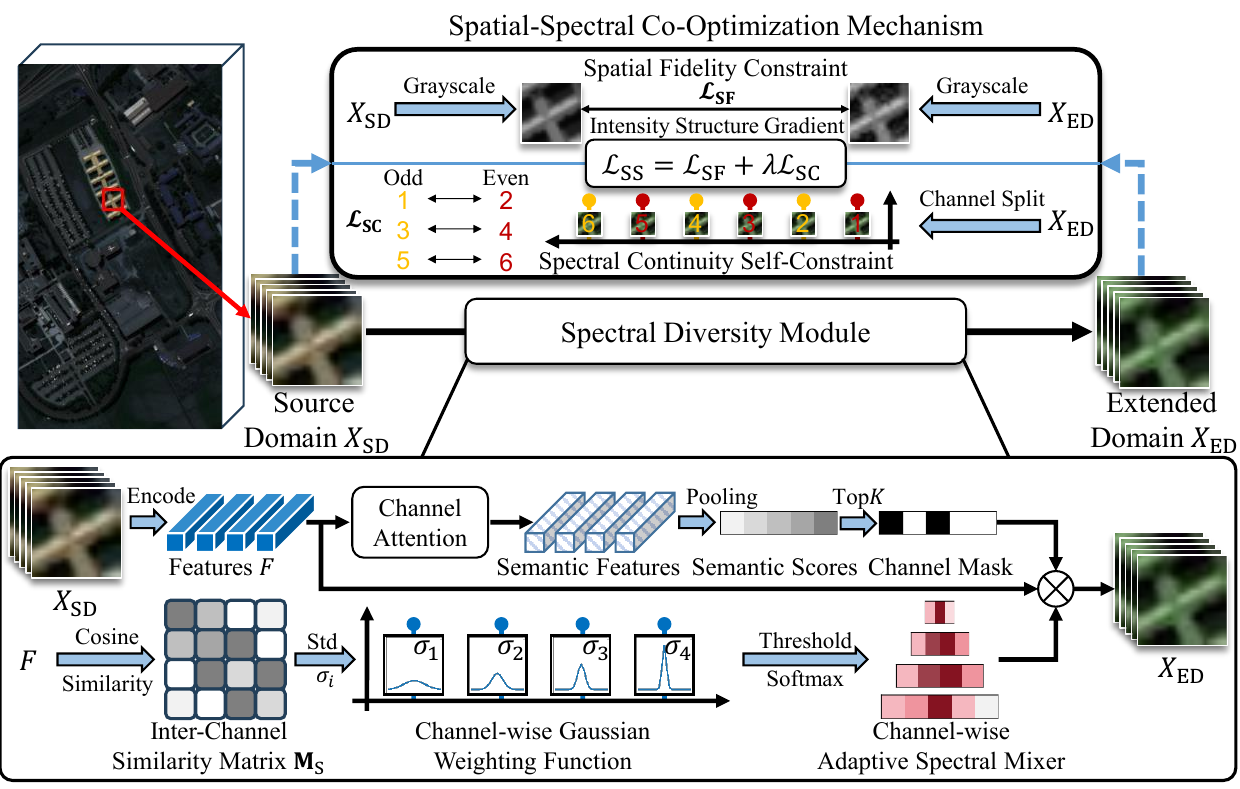}
    \caption{Overview of spectral property-driven data augmentation (SPDDA). Briefly, a spectral diversity module (SDM) is utilized to inject perturbations into source-domain samples $X_\mathrm{SD} \in \mathbb{R}^{H \times W \times C}$ to obtain extended-domain samples $X_\mathrm{ED}\in \mathbb{R}^{H \times W \times K}$, where $K$ is a random integer. SDM constructs a channel mask and a channel-wise adaptive spectral mixer to masks an arbitrary number of channels and mixes the remaining channels, which adheres to the practical phenomenon in hyperspectral image. The augmentation procedure is supervised by a spatial-spectral co-optimization mechanism, which is comprised of a spatial fidelity constraint and a spectral continuity self-constraint.}
    \label{fig:fig1}
\end{figure*}

\section{Methodology}
\subsection{Framework Overview}
Although existing hyperspectral image (HSI) classification methods have achieved excellent performance in remote sensing, they typically assume that the processed samples are independent and identically distributed. Thus, the cross-domain shifts that exist in the practical scenarios would result in a drastic performance degradation. Under the condition of single-source domain samples availability, pioneering works address this issue by generating augmented samples to simulate domain shifts between different domains. However, there is a tradeoff between realism and diversity during the generation process, which constraints their performance. To tackle this challenge, we propose a spectral property-driven data augmentation (SPDDA), which obtains the augmented samples by accounting for inherent properties of HSI. The framework of SPDDA is illustrated in Fig.~\ref{fig:fig1}.

SPDDA develops a spectral diversity module to inject perturbations to the source-domain samples $X_{\mathrm{SD}} \in \mathbb{R}^{H \times W \times C}$. Firstly, $X_{\mathrm{SD}}$ is encoded by multiple ResBlock~\cite{resnet} and obtain features $F \in \mathbb{R}^{H \times W \times C}$, where $H$ and $W$ denote the spatial size, $C$ denotes the number of spectral channels. Then a channel mask and a channel-wise adaptive spectral mixer are constructed based on the semantic information and inter-channel similarity of $F$, respectively. Finally, we integrate channel mask, adaptive spectral mixer and $F$ by channel-wise multiplication to acquire the augmented samples $X_{\mathrm{ED}} \in \mathbb{R}^{H \times W \times K}$, where $K$ is a random integer. To supervise the sample generation process, a spatial-spectral co-optimization mechanism is proposed to enforce $X_\mathrm{ED}$ to adhere to spatial fidelity constraint $\mathcal{L}_\mathrm{SF}$ and spectral continuity self-constraint $\mathcal{L}_\mathrm{SC}$ simultaneously. In addition, the weight of spectral constraint is adjusted adaptively based on $\|\mathcal{L}_\mathrm{SC}\|_1$, thereby achieving a dynamic balance and cooperative optimization between two constraints.

\subsection{Spectral Diversity Module}

A spectral diversity module (SDM) is proposed to simulate device-dependent variations in the number of spectral channels and the mixing of adjacent channels, thereby injecting perturbations for $X_\mathrm{SD}$ and generating diverse extended samples $X_\mathrm{ED}$. The structure of SDM is depicted in Fig.~\ref{fig:fig1}. In brief, SDM constructs a channel mask and a channel-wise adaptive spectral mixer (CASM), where the former is utilized to mask channels with low semantic scores to obtain HSI with an arbitrary number of channel. CASM accounts for the linear spectral mixing model and constructs a channel-wise Gaussian weighting function based on inter-channel similarity to obtain mixing kernels with varying weights and sizes. Details are as follows.


SDM firstly encodes $X_\mathrm{SD}$ to obtain features $F$ through multiple ResBlock. Then a channel-wise self attention mechanism~\cite{vit} is utilized to extract semantic information from $F$, and obtain semantic features $F_\mathrm{s}$. In brief, a layer normalization is first employed to normalize $F$, a convolution group is further utilized to generate query token $\mathbf{Q}$. The key token $\mathbf{K}$ and the value token $\mathbf{V}$ are obtained by the same manner. Then, attention matrix $\mathbf{A} \in \mathbb{R}^{C \times C}$ is acquired through a dot product between $\mathbf{Q}$ and $\mathbf{K}$. Finally, we integrate $\mathbf{A}$ and $\mathbf{V}$ by matrix production to focus on more valuable channels, and a feed-forward network~\cite{FFN} is further employed to enhance representative ability to obtain $F_\mathrm{S}$.

Furthermore, $F_\mathrm{S}$ is compressed by the pooling layers to serve as a semantic score for each channel. The channels with low scores are masked to obtain the channel mask. The calculation is formulated as:
\begin{equation}
    S = \mathrm{Avg}(F_\mathrm{S}) + \mathrm{Max}(F_\mathrm{S}),
\end{equation}
where $S \in \mathbb{R}^{1 \times C}$ denotes the semantic score, $\mathrm{Avg}(\cdot)$ denotes the average pooling, and $\mathrm{Max}(\cdot)$ denotes the maximum pooling. Then, the channel mask is acquired by activating the first $K$ channels and masking the remaining depending on $S$, where $K$ is a random integer.




Moreover, we design the CSAM to enable diverse augmentation patterns. Specifically, we employ cosine similarity~\cite{guan2025ot} to characterize correlation between channels of $F$, and obtain the counterpart matrix $\mathbf{M}_\mathrm{S} \in \mathbb{R}^{C \times C}$. Then, we compute the row-wise standard deviation $\sigma_i$ of $\mathbf{M}_\mathrm{S}$ to characterize the distribution of similarities between each channel and the others, where $i \in [1,\dots, C]$ denotes the $i$-th channel. $\sigma_i$ is further used to act as the scale parameter to construct channel-wise Gaussian weighting function, achieving channel-tailored adaptive adjustment while keeping continuous and differentiable. The above procedure is formulated as follows.
\begin{equation}
    w_j^i = \frac{\mathrm{e}^{-\frac{l(j)^2}{2\sigma^2}}}{\sqrt{2\pi}\sigma_i}, 
    \label{equ_gas}
\end{equation}
where $w_j^i$ denotes the $j$-th weight in the $i$-th mixing kernel, $j \in [-p, p]$, $p$ is a human-set parameter to indicate the maximum mixer size, and $l(j) = m + \epsilon (j+p)$. A thresholding operation is then applied to $w_j^i$, which preserves the elements that satisfy $w_j^i \leq \sigma_i$, and then $w^i$ is normalized via softmax function to complete the construction of CSAM.


After that, feature $F$ is masked by the channel mask to obtain HSI $X_\mathrm{mask}$ with arbitrary number of channels. Then CSAM introduces perturbations in the spectral dimension via channel-wise multiplication, which is formulated as:
\begin{equation}
    X_\mathrm{ED}^i = \sum_{j=-p}^{j=p}X_\mathrm{mask}^{i+j} \times w_j^i,
\end{equation}
where $X_\mathrm{ED}^i$ denotes the $i$-th channel of $X_\mathrm{ED}$, and $X_\mathrm{mask}$ is padded to ensure the correctness of computation.

\subsection{Spatial-Spectral Co-Optimization Mechanism}
\begin{figure}
    \centering
    \includegraphics[width=\linewidth]{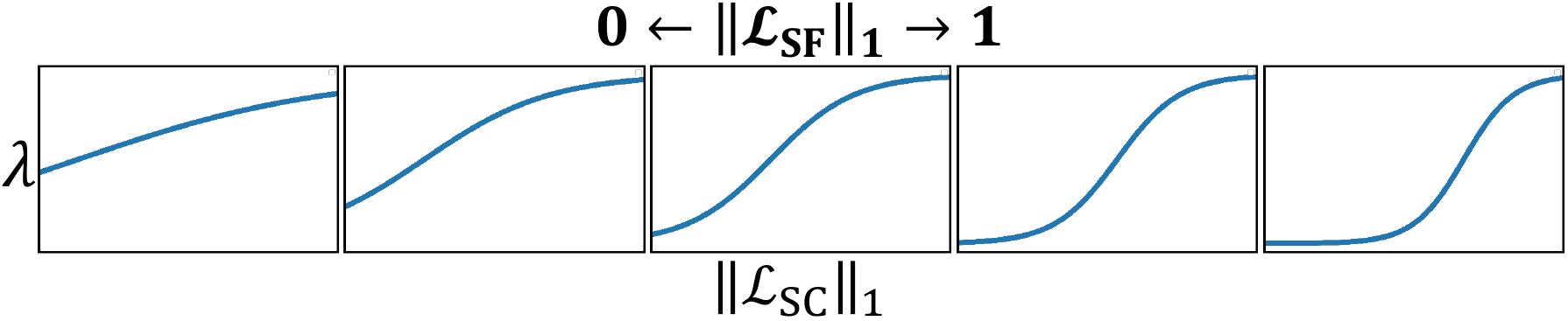}
    \caption{Illustration of the variation in the spectral continuity self-constraint weight $\lambda$.}
    \label{fig:fig2}
\end{figure}
Furthermore, we design a spatial-spectral co-optimization mechanism (SSCOM) to ensure that the spatial information described by $X_\mathrm{ED}$ is consistent with $X_\mathrm{SD}$, while preventing excessive aliasing from disrupting the continuity of the spectral dimension. SSCOM comprises a spatial fidelity constraint $\mathcal{L}_\mathrm{SF}$ and a spectral continuity self-constraint $\mathcal{L}_\mathrm{SC}$, which is formulated as:
\begin{equation}
    \mathcal{L}_\mathrm{SS} = \mathcal{L}_\mathrm{SF} + \lambda \mathcal{L}_\mathrm{SC},
    \label{equ_SSCOM}
\end{equation}
where $\mathcal{L}_\mathrm{SS}$ denotes the optimized objective of SSCOM, $\lambda$ denotes an adaptive coefficient, aiming to dynamically adjust the weight of $\mathcal{L}_\mathrm{SC}$ in accordance with the optimization progress of $\mathcal{L}_\mathrm{SF}$.

For the spatial fidelity constraint, we grayscale $X_\mathrm{ED}$ and $X_\mathrm{SD}$, and denote counterparts as $I_\mathrm{ED}$ and $I_\mathrm{SD}$, respectively. Then gray images from different domains are forced to share similar spatial information such as structure and texture, which ensures the spectral diversity of the augmented samples is not constrained by the source domain. Accordingly, we formulate $\mathcal{L}_\mathrm{SF}$ as:
\begin{equation}
    \begin{split}
        \mathcal{L}_\mathrm{SF} = \mathcal{L}_\mathrm{MSE}(I_\mathrm{SD}, I_\mathrm{ED}) + \mathcal{L}_\mathrm{SSIM}(I_\mathrm{SD}, I_\mathrm{ED}) \\
        + \mathcal{L}_\mathrm{GRA}(I_\mathrm{SD}, I_\mathrm{ED}),
    \end{split}
    \label{equ_SF}
\end{equation}
where $\mathcal{L}_\mathrm{MSE}(\cdot, \cdot)$ denotes the mean square error (MSE) loss, $\mathcal{L}_\mathrm{SSIM}(\cdot, \cdot)$ denotes the structural similarity index measure (SSIM) loss, and $\mathcal{L}_\mathrm{GRA}(\cdot, \cdot)$ denotes the gradient loss. 

In addition, we introduce the spectral continuity self-constraint by taking into account the inherent property of HSI, i.e., neighboring channels characterize similar content. Briefly, we split $X_\mathrm{ED}$ by odd-even channel, which are denoted as $X_\mathrm{ED}^\mathrm{E}$ and $X_\mathrm{ED}^\mathrm{O}$. Then they are constrained by MSE and SSIM, which is formulated as:
\begin{equation}
    \mathcal{L}_\mathrm{SC} = \mathcal{L}_\mathrm{MSE}(X_\mathrm{ED}^\mathrm{E}, X_\mathrm{ED}^\mathrm{O}) + \mathcal{L}_\mathrm{SSIM}(X_\mathrm{ED}^\mathrm{E}, X_\mathrm{ED}^\mathrm{O}).
    \label{equ_SC}
\end{equation}

\begin{table*}[htbp]
\centering
\begin{tabular}{*{10}{c}}

  \toprule
  \multirow{2}*{Methods} & \multicolumn{3}{c}{Houston} & \multicolumn{3}{c}{Pavia} & \multicolumn{3}{c}{HyRank} \\  
  \cmidrule(lr){2-4}\cmidrule(lr){5-7}\cmidrule(lr){8-10}
  & OA $\uparrow$ & F1 $\uparrow$ & Kappa $\uparrow$ & OA $\uparrow$ & F1 $\uparrow$ & Kappa $\uparrow$ & OA $\uparrow$ & F1 $\uparrow$ & Kappa $\uparrow$ \\
  \midrule
  SDENet & 67.02 & 0.6768 & 0.4702 & \underline{79.05} & \underline{0.7860} & \underline{0.7471} & 58.40 & \underline{0.5574} & 0.4916 \\
  FDGNet & 75.44 & 0.7329 & 0.5360 & 78.28 & 0.7736 & 0.7371 & 56.81 & 0.5035 & 0.4554  \\
  S2AMSNet & 71.13 & 0.7291 & 0.5541 & 73.42 & 0.7303 & 0.6792 & 59.46 & 0.5561 & 0.4940 \\
  S2ECNet & 72.84 & 0.7355 & 0.5641  & 78.05 & 0.7828 & 0.7358 & \underline{60.21} & \textbf{0.5617} & \textbf{0.5047}\\
  ProRandConv & \underline{75.71} & \underline{0.7635} & \underline{0.5911} & 78.27 & 0.7763 & 0.7375 & 57.76 & 0.5295 & 0.4753\\
  ABA & 52.68	& 0.5634 & 0.3321 & 57.87 & 0.5698 & 0.5070 & 48.31 & 0.3625 & 0.3235  \\
  StyDeSty & 70.99 & 0.7155 & 0.5260 & 70.33 & 0.7075 & 0.6432 & 55.25 & 0.4559 & 0.4255 \\
  Ours & \textbf{78.60} & \textbf{0.7826} & \textbf{0.6252} & \textbf{83.03} & \textbf{0.8348} & \textbf{0.7963} & \textbf{60.70} & 0.5316 & \underline{0.5024}\\
  \bottomrule
\end{tabular}
\caption{Quantitative Results on Cross-scene Hyperspectral Image Classification for Remote Sensing. The optimal results are bold, and the sub-optimal results are underlined.}
\label{tab_1}
\end{table*}

Although $\mathcal{L}_\mathrm{SC}$ facilitates maintaining spectral continuity, it lacks effective supervisory information. When the network places excessive focus on this term during training, it may lead to augmented samples that are overly smoothed in the spectral dimension, neglecting spatial structure preservation and thereby reducing both diversity and realism. To tackle this issue, we implement a nonlinear adjustment of the weight for $\mathcal{L}_\mathrm{SC}$ based on the hyperbolic tangent function and employ $\|\mathcal{L}_\mathrm{SF}\|_1$ to dynamically tune the translation and curvature change of the function. Hence, the decreasing rate of $\lambda$ accelerates and enables reaching a lower value when $\|\mathcal{L}_\mathrm{SF}\|_1$ takes a higher value, whose variation is illustrated in Fig.~\ref{fig:fig2}. The mentioned procedure is formulated as: 
\begin{equation}
\left\{
\begin{aligned}
\lambda &= 1 + \frac{\mathrm{e}^t - \mathrm{e}^{-t}}{\mathrm{e}^t + \mathrm{e}^{-t}}, \\
t &= \frac{\|\mathcal{L}_\mathrm{SC}\|_1 - f_\mathrm{T}(\|\mathcal{L}_\mathrm{SF}\|_1)}{f_\mathrm{C}(\|\mathcal{L}_\mathrm{SF}\|_1)},
\end{aligned}
\right.
\label{equ_lam}
\end{equation}
where $f_\mathrm{T}(\cdot)=\frac{s\|\mathcal{L}_\mathrm{SF}\|_1}{2}$ denotes the translating function, $f_\mathrm{C}(\cdot)=\frac{1}{s\|\mathcal{L}_\mathrm{SF}\|_1}$ denotes the curvature function, $s$ denotes a scaling parameter. 



\subsection{Loss Functions}
Except for SSCOM $\mathcal{L}_\mathrm{SS}$, a cross-entropy loss $\mathcal{L}_\mathrm{CE}$ is also employed to supervise the training procedure of SPDDA, whose optimized objective is formulated as:
\begin{equation}
    \mathcal{L}_\mathrm{total} = \mathcal{L}_\mathrm{SS} + \mathcal{L}_\mathrm{CE}(P, y),
\end{equation}
where $P$ denotes the prediction outputted by the classifier which takes $X_\mathrm{ED}$ as inputs, $y$ denotes the category label.

\section{Experiments and Results}
\subsection{Experimental Settings}

\subsubsection{Datasets}
We employ three cross-scene hyperspectral image classification datasets~\cite{dataset_1, dataset_2} to evaluate the proposed method, including Houston, Pavia, and HyRank. Both datasets contain two scenes, which were acquired by capturing different areas in the same time or capturing the same areas in different years using distinct imaging devices. For Houston, Houston13 and Houston18 are used as the source domain (SD) and target domain (TD), respectively. In the case of Pavia, PaviaU serves as the SD and PaviaC as the TD. For HyRank, Dioni and Loukia are designated as the SD and TD, respectively. Their number of categories is 7, 7, and 12, respectively. In addition, each dataset is partitioned into $13 \times 13$ patches, where the label of the central pixel is assigned as the category of the corresponding patch.





\subsubsection{Implementation Details}
In all experiments, we employ LeNet~\cite{lenet} as the classifier, where the first convolution is replaced by a channel-adaptive convolution~\cite{ada_conv} to suit inputs with varying numbers of channels. During training, the chosen optimizer is Adam with the learning rate $1\times10^{-4}$. The hyperparameters in Eq.~\ref{equ_gas} is set as: $m=-5$, $\epsilon=0.5$, and $p=10$. The hyperparameter in Eq.~\ref{equ_lam} is set as: $s=15$. Following existing methods~\cite{SDENet}, we linearly integrate SD and ED to generate intermediate-domain samples and train the classifier using both cross-entropy loss and contrastive loss~\cite{khosla2020supervised}. The classifier of each comparative method is kept the same as the proposed method. All experiments are conducted on the Pytorch platform, and the computing device is equipped with an NVIDIA RTX 4090D GPU. Each method is trained for $400$ epochs. 

\subsubsection{Evaluation Metrics}
Several evaluation metrics are employed to quantify results of each method, including the overall accuracy (OA), the weighted F1 score (F1), and the kappa coefficient (Kappa).

\subsection{Comparative Results}
To demonstrate the superiority of our method, we employ multiple state-of-the-art (SOTA) methods in hyperspectral single-source domain generalization (SDG) as comparative methods, comprising of SDENet~\cite{SDENet}, FDGNet~\cite{FDGNet}, S2ECNet~\cite{S2ECNet}, and S2AMSNet~\cite{S2AMSnet}. In addition, several SOTA methods for RGB images also taken as comparison, including ProRandConv~\cite{prorandconv}, ABA~\cite{ABA}, and StyDeSty~\cite{StyDeSty}. To ensure a fair comparison, all the above methods are implemented using their official open-source codes, with configurations strictly following those reported in their respective papers.

\begin{figure*}
    \centering
    \includegraphics[width=\linewidth]{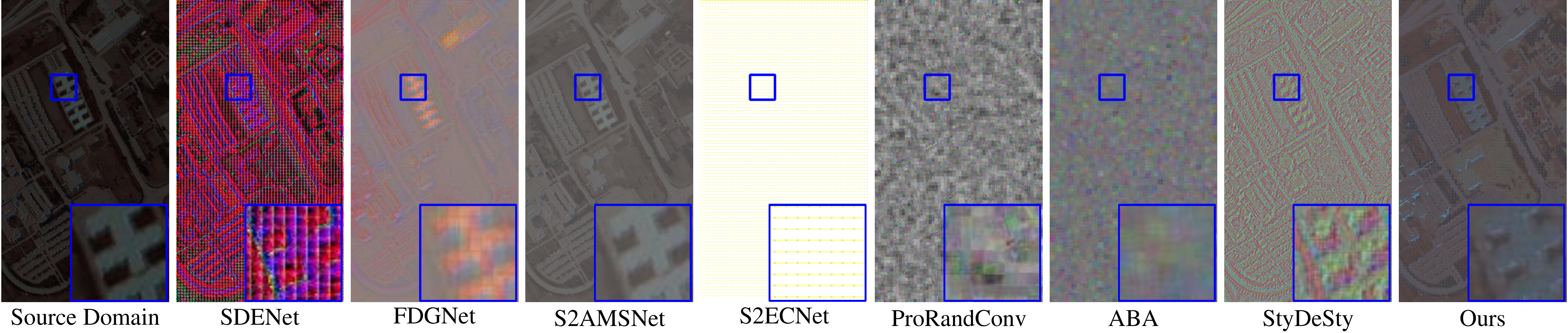}
    \caption{Illustration of the pseudo-color images for the extended-domain scene for each method. Regions marked by the blue rectangle are enlarged and displayed in the lower right corner of each image.}
    \label{fig:vis}
\end{figure*}

\subsubsection{Quantitative Analysis} Experimental datasets are configured as Houston, Pavia, and HyRank. Quantitative results are presented in Tab.~\ref{tab_1}. In terms of OA, the proposed method achieves optimal results compared to the comparisons, indicating the significant advantages of our method in hyperspectral SDG. FDGNet introduces geometry constraints to ensure that the augmented samples keep realism, resulting in a unbiased distribution compared with other comparative methods. However, such constraints often suppress the diversity of augmented samples, thereby limiting the generalization ability of the classifier. A similar issue is even more pronounced in S2AMSNet. In addition, the RGB-based method ProRandConv employs random convolution layers to perturb the local semantic information of source-domain samples, thereby obtaining diverse augmented samples and introducing extra spatial information like structure for the classifier. However, blind convolution may disrupt spectral characteristics and thus impair classification performance. In contrast to existing methods, the proposed method generates augmented samples by accounting for the inherent spectral properties, i.e., simulating real-world variations in spectral channel counts and aliasing between adjacent channels. Due to this generative manner, the proposed method is able to preserve both the realism and diversity of the augmented samples, thereby achieving superior performance.

\subsubsection{Qualitative Analysis}
\begin{figure}
    \centering
    \includegraphics[width=\linewidth]{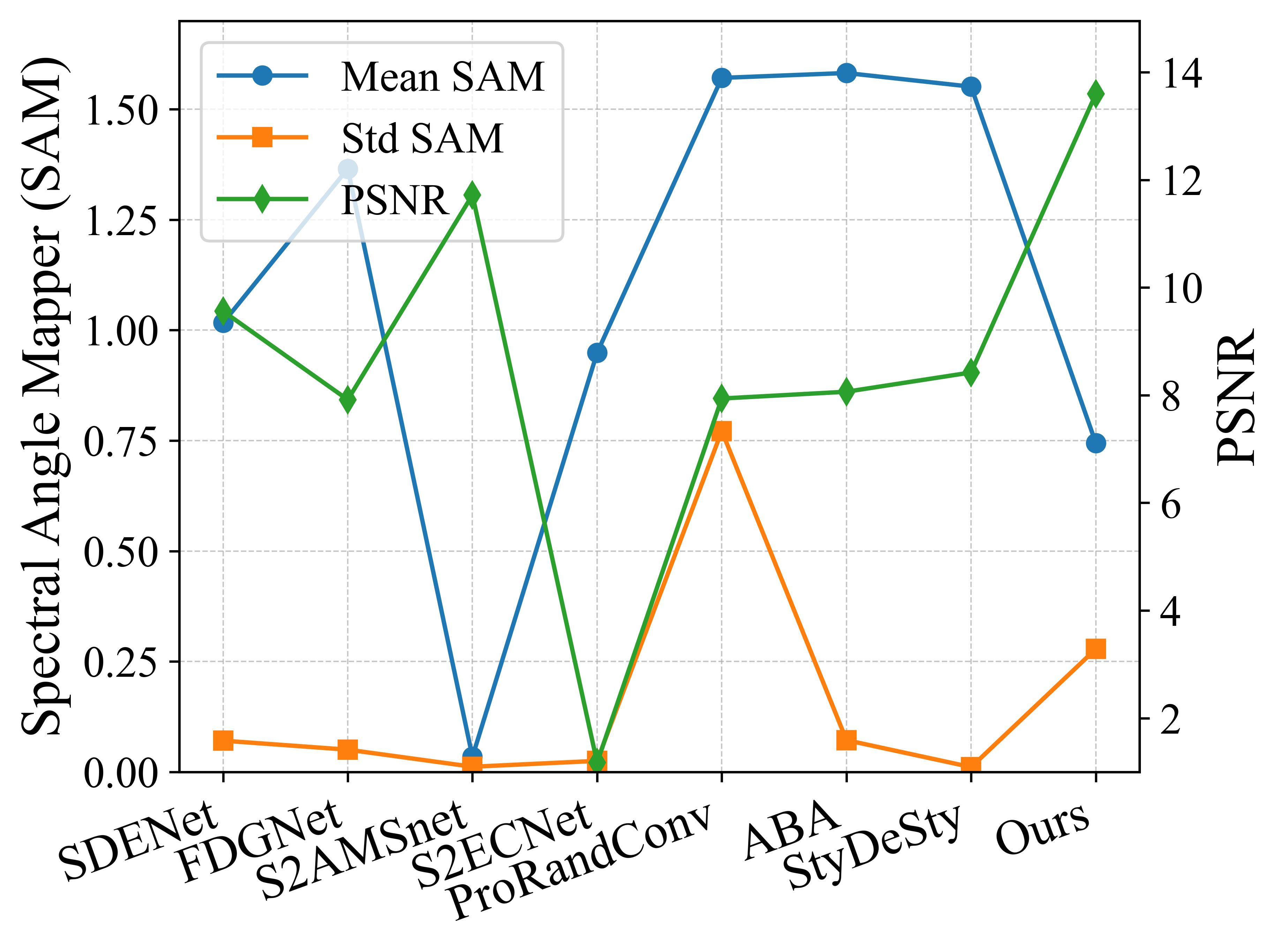}
    \caption{Comparison of realism and diversity. PSNR measures spatial realism, while the mean and standard deviation (Std) of SAM represent spectral realism and diversity, respectively. A lower mean SAM indicates higher spectral realism, and a greater Std means higher diversity.}
    \label{fig:div}
\end{figure}

We further conduct a qualitative evaluation of the proposed method and the comparison methods on the Pavia dataset. Firstly, all generated samples are sequentially integrated to construct the entire extended-domain scene, and pseudo-color images are produced for qualitative visualization, which is illustrated in Fig.~\ref{fig:vis}. To facilitate a more effective comparison, we enlarge regions highlighted by the blue rectangle in each visualization. In addition, we evaluate the peak signal to noise ratio (PSNR) of the pseudo-color images to quantify the spatial realism for each method, taking the source-domain counterpart as the reference. The spectral angle mapper (SAM)~\cite{SAM} is also computed, where SAM assesses the similarity between source-domain samples and generated data based on the angular difference in spectral dimension. The mean value and the standard deviation of SAM are employed to quantify the realism and diversity, respectively. The comparison is depicted in Fig.~\ref{fig:div}.

ProRandConv and ABA severely compromise the semantic consistency of the samples due to the utilization of convolution-based augmentation strategy. Although StyDeSty explicitly optimizes the reliability of generated samples during training, it still introduces certain distortions as it considers only spatial content. S2ECNet concentrates exclusively on domain-invariant feature extraction during training, resulting in a tendency to produce meaningless samples, which is shown in Fig.~\ref{fig:vis}. In comparison, samples produced by the hyperspectral image-tailored S2AMSNet exhibit superior realism. However, their diversity is limited due to the introduction of strict constraints based on the source-domain samples, which is depicted in Fig.~\ref{fig:div}. In contrast to comparisons, the proposed method merely establishes basic grayscale constraints with the original samples and flexibly injects perturbations in accordance with inherent heterogeneity and mixing phenomena in the spectral dimension. Therefore, our method effectively addresses the tradeoff between realism and diversity.


\subsection{Ablation Study}

\subsubsection{Module Ablation Experiments}
\begin{figure}
    \centering
    \includegraphics[width=\linewidth]{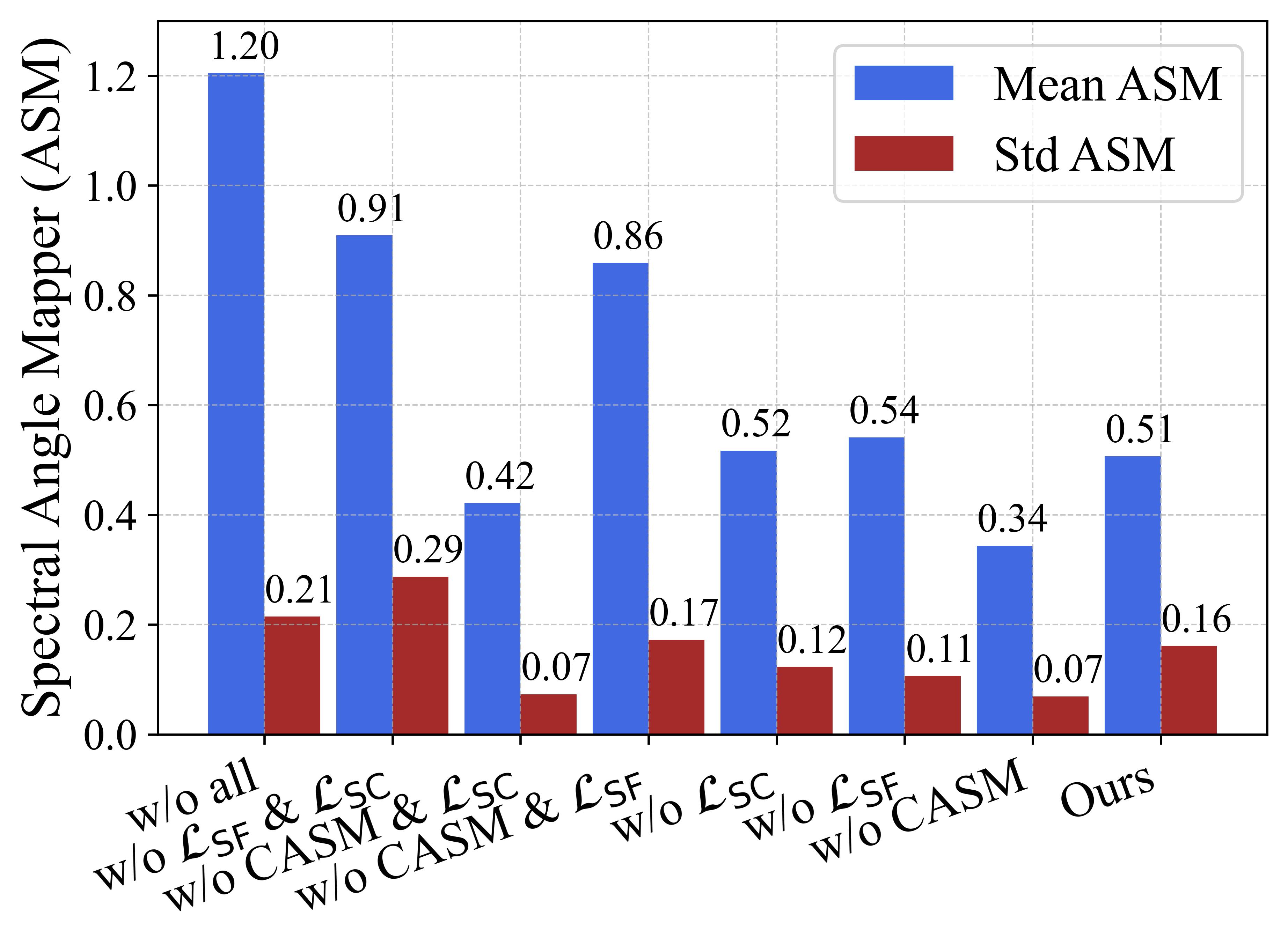}
    \caption{Effects of CASM, $\mathcal{L}_\mathrm{SF}$, and $\mathcal{L}_\mathrm{SC}$ on spectral angle mapper (SAM). The average SAM quantifies realism of generation. A lower average SAM indicates higher realism.}
    \label{fig:sam}
\end{figure}

We then conduct ablation experiments on Houston to verify the effect of channel-wise adaptive spectral mixer (CASM), spatial fidelity constraint ($\mathcal{L}_\mathrm{SF}$), and spectral continuity self-constraint ($\mathcal{L}_\mathrm{SC}$). In addition, we employ a fixed mixer as a substitute for CASM. Results are presented in Tab.~\ref{tab2}. Except for the accuracy metrics, we also compute the mean SAM and Std SAM to quantify realism and diversity in the spectral dimension, respectively. Results are illustrated in Fig.~\ref{fig:sam}.

\begin{table}[htbp]
  \centering
  
  \begin{tabular}{c c c c c c}
    \toprule
    CASM & $\mathcal{L}_\mathrm{SF}$ & $\mathcal{L}_\mathrm{SC}$ & OA $\uparrow$ & F1 $\uparrow$ & Kappa $\uparrow$\\ 
    \midrule
    $\times$ & $\times$ & $\times$ & 74.73 & {0.7637} & 0.6068  \\
    $\checkmark$ & $\times$ & $\times$ & 76.95 & \underline{0.7794} & \textbf{0.6264}  \\
    $\times$ & $\checkmark$ & $\times$ & 70.93 & 0.7252 & 0.5408 \\
    $\times$ & $\times$ & $\checkmark$ & 72.23 & 0.7414 & 0.5711  \\
    $\checkmark$ & $\checkmark$ & $\times$ & \underline{77.87} & 0.7724 & 0.6081\\
    $\checkmark$ & $\times$ & $\checkmark$ & 75.94 & 0.7643 & 0.6003  \\
    $\times$ & $\checkmark$ & $\checkmark$ & 71.26 & 0.7299 & 0.5433 \\
    $\checkmark$ & $\checkmark$ & $\checkmark$ & \textbf{78.60} & \textbf{0.7826} & \underline{0.6252}\\
    \bottomrule
    \end{tabular}
    \caption{Ablation study of channel-wise adaptive spectral mixer (CASM), spatial fidelity constraint ($\mathcal{L}_\mathrm{SF}$), and spectral continuity self-constraint ($\mathcal{L}_\mathrm{SC}$) on Houston.}
    
    \label{tab2}
\end{table}

Pairing the model solely with CASM can effectively enhance realism and diversity, thereby facilitating improvements in classifier performance in the unseen target domain. Conversely, $\mathcal{L}_\mathrm{SF}$ or $\mathcal{L}_\mathrm{SC}$ induces a drastic decline in classification accuracy. As seen in Fig.~\ref{fig:sam}, this degradation likely stems from the fixed mixing pattern suppressing the diversity of generated samples, which is further aggravated by the introduction of $\mathcal{L}_\mathrm{SF}$ and $\mathcal{L}_\mathrm{SC}$. Compared to $\mathcal{L}_\mathrm{SC}$, $\mathcal{L}_\mathrm{SF}$ enforces augmented samples are similar to source-domain samples in the grayscale domain, causing more severe diversity loss. On the other hand, it improves the realism of sample generation, thus mitigating the adverse impact of unrealistic samples on classification performance. Therefore, the classifier performance can be effectively improved when $\mathcal{L}_\mathrm{SF}$ is combined with CASM. When three components are combined, the tradeoff between diversity and realism is further alleviated, and the performance achieves optimal.


\subsubsection{$\lambda$-tuning Ablation Experiments}
We further conduct $\lambda$-tuning ($\lambda$-T) experiments to evaluate the effects of the adaptive tuning mechanism described in Eq.~\ref{equ_lam}. Specifically, we modify the tuning strategy of $\lambda$ and obtained seven variants. In the first five versions, $\lambda$ is set to a fixed value, i.e., $\lambda \in \{0.1, 0.5, 1.0, 1.5, 2.0\}$, which are respectively denoted as Fix-0.1, Fix-0.5, etc. In addition, the last two versions divide the training process into early and late stages, which are denoted as $\mathrm{Var}_\mathrm{E}$ and $\mathrm{Var}_\mathrm{S}$, respectively. In the early stage, $\lambda$ is configured to $0.1$ and to $2.0$ during the late stage, where $0.1$ is close to the lower bound of Eq.~\ref{equ_lam}, $2.0$ is the upper bound of Eq.~\ref{equ_lam}. $\mathrm{Var}_\mathrm{E}$ defines the early training stage as epochs less than $200$, whereas $\mathrm{Var}_\mathrm{S}$ considers the training to be in the early stage when $\|\mathcal{L}_\mathrm{SF}\|_1 > 0.08$, which is determined based on the empirical observations.

  
    

\begin{table}[htbp]
  \centering
  
  \begin{tabular}{*{4}{c}}
    \toprule
    $\lambda$-T & OA $\uparrow$ & F1 $\uparrow$ & Kappa $\uparrow$ \\ 
    \midrule
    Fix-0.1 & 75.19 & 0.7563 & 0.5677  \\
    Fix-0.5 & 75.18 & 0.7651 & 0.5877 \\
    Fix-1.0 & 72.05 & 0.7398 & 0.5374 \\
    Fix-1.5 & 75.27 & 0.7662 & 0.5800 \\
    Fix-2.0 & 74.47 & 0.7620 & 0.5729 \\
    $\mathrm{Var}_\mathrm{E}$  & \underline{76.15} & {0.7632} & \underline{0.5906} \\
    $\mathrm{Var}_\mathrm{S}$  & 75.62 & \underline{0.7687} & 0.5817  \\
    Ours & \textbf{78.60} & \textbf{0.7826} & \textbf{0.6252}\\
    \bottomrule
    \end{tabular}
    \caption{Ablation study of $\lambda$-tuning ($\lambda$-T) strategy on Houston. Fix-0.1 denotes that $\lambda$ is set to a fixed value 0.1. $\mathrm{Var}_\mathrm{E}$ and $\mathrm{Var}_\mathrm{S}$ represent early–late stage division based on training epochs and $\|\mathcal{L}_\mathrm{SF}\|_1$, respectively, with $\lambda$ set to 0.1 in the early stage and 2.0 in the late stage.}
    \label{tab3}
\end{table}

The experiments are conducted on the Houston dataset, whose results are presented in Tab.~\ref{tab3}. The fixed $\lambda$ significantly degrades the performance of classifier in the target domain. This may be attributed to the use of fixed weights, which hinder the generator from balancing the spatial fidelity constraint $\mathcal{L}_\mathrm{SF}$ and the spectral continuity self-constraint $\mathcal{L}_\mathrm{SC}$ during training. Specifically, $\mathcal{L}_\mathrm{SF}$ serves as an anchor during the optimization process of generator since it establish a basic grayscale constraint with source-domain samples. $\mathcal{L}_\mathrm{SC}$ serves as a regularization term through maintaining inter-channel continuity. Therefore, as training progresses, the optimization focus of generator should gradually shifts between the two constraints. Accordingly, $\mathrm{Var}_\mathrm{E}$ and $\mathrm{Var}_\mathrm{S}$ achieve this objective by employing a two-stage adjustment strategy, which prioritizes spatial fidelity during the early training stage and focuses on optimizing spectral continuity in the later stage. However, it remains challenging to establish a criterion for dividing these stages to achieve satisfactory results. In contrast, the proposed adaptive tuning strategy adjusts $\lambda$ based on $\|\mathcal{L}_\mathrm{SF}\|_1$, effectively enabling generator to shift its optimization focus between the two constraints adaptively and achieve optimal performance.

\section{Conclusion}
In this paper, we propose a spectral property-driven data augmentation (SPDDA) to mitigate the dilemma that the tradeoff between realism and diversity of the augmented samples in hyperspectral single-source domain generalization. Specifically, SPDDA develops a spectral diversity module (SDM) to simulate the variation in  spectral channel counts and the spectral mixing among adjacent channels, which are inherent phenomena in real-world scenarios. To achieve diverse mixing patterns, SDM constructs a channel-wise adaptive spectral mixer through modeling a channel-wise Gaussian weighting function based on the inter-channel similarity. Moreover, we propose a spatial-spectral co-optimization mechanism (SSCOM) that comprises a spatial fidelity constraint $\mathcal{L}_\mathrm{SF}$ and a spectral continuity self-constraint $\mathcal{L}_\mathrm{SC}$ to ensure that generation maintains realism. The weight of $\mathcal{L}_\mathrm{SC}$ is further adjusted adaptively according to $\mathcal{L}_\mathrm{SF}$, realizing a dynamical balance between two constraints during training. The effectiveness of the proposed method is identified by comprehensive experiments on three open-source datasets.

\section{Acknowledgments}
This work was supported in part by the National Natural Science Foundation of China under Grant 62031023 and 62331011, and in part by Shenzhen Science and Technology Project under Grant GXWD20220818170353009.

\bibliography{aaai2026}

\end{document}